\documentclass[twocolumn,natbib]{svjour3}          %
\smartqed  %

\usepackage{tabularx}
\usepackage{subfiles}
\usepackage{arydshln}
\usepackage{xcolor}
\usepackage{etoolbox}
\usepackage{subcaption}
\usepackage{natbib}
\usepackage{graphicx}
\usepackage{makecell}
\usepackage[group-separator={,},output-decimal-marker = {.}, group-minimum-digits={3}]{siunitx}  %
\usepackage[pagebackref=true,breaklinks=true,colorlinks,bookmarks=false,draft]{hyperref}

\usepackage[inline]{enumitem}
\usepackage{enumitem}%

\usepackage{ marvosym }
\providetoggle{showcomments}									%
\settoggle{showcomments}{false} 								%

\providetoggle{longversion}										%
\settoggle{longversion}{true} 								%

\renewcommand{\cite}{\citep}
\newcommand{\para}[1]{\vspace{4pt}\noindent\textbf{#1}}

\newcommand{\eg}{\textit{e.g.} }
\newcommand{\etc}{\textit{etc.} }
\newcommand{\ie}{\textit{i.e.} }
\newcommand{\vs}{\textit{vs.} }
\definecolor{brown}{rgb}{0.59, 0.29, 0.0}
\definecolor{darkgreen}{rgb}{0, 0.569, 0}
\definecolor{darkyellow}{rgb}{0.9, 0.9, 0.2}
\definecolor{darkred}{rgb}{0.4, 0, 0}
\definecolor{darkgray}{rgb}{0.3, 0.3, 0.3}

\newcommand{\changedtext}[1]{{\color{black}#1}}

\newcommand{\attc}[1]{\textcolor{black}{{#1}}} %
\newcommand{\cam}[1]{#1}
\iftoggle{showcomments}{%
	\newcommand{\michael}[1]{\textcolor{blue}{{ Michael: #1}}}
	\newcommand{\vitto}[1]{\textcolor{red}{{ Vitto: #1}}}
	\newcommand{\jordi}[1]{\textcolor{purple}{{ Jordi: #1}}}
	\newcommand{\old}[1]{\color{brown}{ OLD: #1 :ENDOLD }\color{black}}
	\newcommand{\att}[1]{\textcolor{black}{{#1}}} %
	\newcommand{\todo}[1]{\textcolor{blue}{{ TODO: #1}}}
	\newcommand{\done}{\michael{DONE}}

}{%

	\newcommand{\michael}[1]{\textcolor{blue}{{}}}
	\newcommand{\done}[1]{}
	\newcommand{\vitto}[1]{\textcolor{red}{{}}}
	\newcommand{\jordi}[1]{\textcolor{purple}{{}}}
	\newcommand{\old}[1]{}
	\newcommand{\att}[1]{\textcolor{black}{{#1}}} %
	\newcommand{\todo}[1]{\textcolor{blue}{{}}}
}

\newcolumntype{H}{>{\setbox0=\hbox\bgroup}c<{\egroup}@{}}
\newcommand{\coco}{COCO}
\newcommand{\ilsvrc}{ILSVRC}

\newcommand{\taskname}{object class labelling}
\newcommand{\jointtask}{class and box labelling}
\renewcommand{\emph}[1]{\textit{#1}}

\journalname{International Journal of Computer Vision}

\begin{document}

\title{Efficient Object Annotation via Speaking and Pointing
}

\author{Michael~Gygli \and
Vittorio~Ferrari}

\institute{M. Gygli \Letter \at
Google Research \\
\email{gyglim@google.com}           %
\and
V. Ferrari \at
Google Research \\
\email{vittoferrari@google.com}           %
}

\date{\vspace{-40pt}}

\maketitle

\begin{abstract}
	Deep neural networks deliver state-of-the-art visual recognition, but they rely on large datasets, which are time-consuming to annotate.
	These datasets are typically annotated in two stages:
	(1) determining the presence of object classes at the image level and
	(2) marking the spatial extent for all objects of these classes.
	In this work we use speech, together with mouse inputs, to speed up this process.
	We first improve stage one, by letting annotators indicate object class presence via speech.
	We then combine the two stages: annotators draw an object bounding box via the mouse and simultaneously provide its class label via speech.
	Using speech has distinct advantages over relying on mouse inputs alone.
	First, it is fast and allows for direct access to the class name, by simply saying it. %
	Second, annotators can simultaneously speak and mark an object location.
	Finally, speech-based interfaces can be kept extremely simple, hence using them requires less mouse movement compared to existing approaches.
	Through extensive experiments on the COCO and ILSVRC datasets we show that our approach yields high-quality annotations at significant speed gains.
	Stage one takes \att{$2.3{\times}-14.9{\times}$} less annotation time than existing methods based on a hierarchical organization of the classes to be annotated.
	Moreover, when combining the two stages, we find that object class labels come for free: annotating them at the same time as bounding boxes has zero additional cost.
	On COCO, this makes the overall process \att{$1.9\times$} faster than the two-stage approach.

\end{abstract}


\section{Introduction}
\label{sec:introduction}
\begin{figure}[t]
  \centering\includegraphics[trim={0 0 0 2cm},clip,width=1\linewidth]{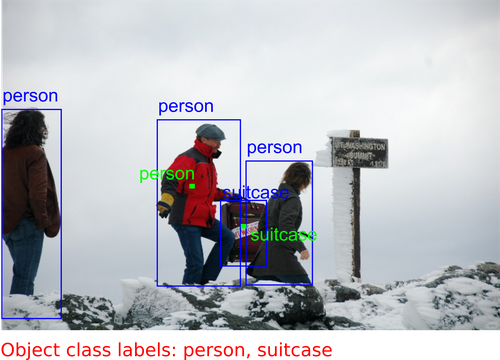}
  \caption{Illustration of common stages of image annotation: annotators first provide object class labels at the image level \cite{russakovsky15ijcv,kuznetsova18arxiv}
  (\textcolor{red}{red}), sometimes associated to a specific object via a click as in \cite{lin14eccv} and our approach for \taskname{} (\textcolor{green}{green}).
  Following stages then annotate the spatial extent of all objects of these classes, \eg with bounding boxes or segmentations (\textcolor{blue}{blue}).
  Speech provides a natural way to combine the two stages to simultaneously annotate class labels and bounding boxes.}
  \label{fig:task_illustration}
\end{figure}

Deep neural networks need millions of training examples to obtain high performance.
Large and diverse datasets such as \ilsvrc{} \cite{russakovsky15ijcv}, COCO \cite{lin14eccv} or Open Images \cite{kuznetsova18arxiv} therefore lie at the heart of the breakthrough and ongoing advances in visual recognition.

Datasets for recognition are typically annotated in two stages \cite{russakovsky15ijcv,kuznetsova18arxiv,lin14eccv,su12aaai} (Fig.~\ref{fig:task_illustration}):
(i) determining the presence or absence of object classes in each image, and
(ii) providing bounding boxes or segmentation masks for all objects of the classes present.
Designing interfaces for stage one, which we call \emph{\taskname{}}, has traditionally been challenging and their use time-consuming.
The key question is how to quickly navigate a vocabulary to find the right classes to annotate.
A na\"ive approach is to ask a separate yes/no question for each class in a given vocabulary.
Such a protocol is rooted on the vocabulary, not the image content. It scales linearly in the size of the vocabulary, even when only few of the classes are present in the image (which is the typical case).
Thus, it becomes very inefficient when the vocabulary is large.
Let us take the \ilsvrc{} dataset as an example:
getting labels for the 200 object classes in its vocabulary would take close to 6 minutes per image~\cite{krishna16chi},
despite each image containing only 1.6 classes on average.
Previous methods have attempted to improve on this by using a hierarchical representation of the class vocabulary to quickly reject certain groups of labels~\cite{lin14eccv,deng14chi}.
This reduces the annotation complexity to sub-linear in the vocabulary size.
But even with these sophisticated methods, \taskname{} remains time consuming.
Using the hierarchical method of~\cite{deng14chi} to label the 200 classes of~\ilsvrc{} still takes 3 minutes per image~\cite{russakovsky15cvpr}.

The COCO dataset has fewer classes (80) and was labelled using the more efficient hierarchical method of~\cite{lin14eccv}. Even so, it still took half a minute per image.

In total, it took \num{20000} hours to annotate object class labels for the COCO dataset \cite{lin14eccv}.
Annotating bounding boxes for these classes (stage two) additionally takes at least \num{5000} hours, even when using efficient box drawing interfaces \cite{papadopoulos17iccv, kuznetsova18arxiv}.
Moreover, the two stages cannot be easily merged to increase efficiency, due to the complexity %
of hierarchical methods.
Such a combined \jointtask{} stage would anyhow be sequential, as annotators cannot simultaneously use the mouse and keyboard to mark the location and provide the class name.

Thus, despite the recent advances in object class label and bounding box annotation, annotating large datasets still requires tremendous amounts of time.
At the same time, \cite{sun17iccv} showed that performance of current deep neural networks is not saturated. These models still benefit from more data, which motivates the community to collect and annotate even larger datasets.

In this paper we propose to use speech, together with mouse pointing, to aid the annotation of such datasets.
First, we use speech for \taskname{} and show that it enables significant speed gains (Sec.~\ref{sec:object_class_labelling}). %
Then, we show that speech allows to naturally combine class and box labelling into one task: annotators mark an object location via the mouse and provide its class label via speech at the same time (Sec. \ref{sec:xclick_task}).
This simultaneous \emph{\jointtask{}} allows to annotate class labels at zero additional cost, compared to annotating bounding boxes alone.

Annotating images via speaking and pointing has multiple strong advantages:
(i) it leads to significant speed gains, as saying the class names is fast: people can say 150 words per minute when describing images \cite{vaidyanathan18acl}.
In comparison, people normally type at 30-100 words per minute \cite{karat99sigchi,clarkson05chi}.
(ii) speaking allows for direct access to the class name via simply \textit{saying it}.
Thereby annotators label classes of objects that they see, ~\ie the task is rooted on the image content and naturally scales with the number of annotated objects.
(iii) it does not require the experiment designer to construct a natural and intuitive hierarchy to access the class labels, as in~\cite{lin14eccv,deng14chi}.
(iv) speaking and pointing can be done in parallel \cite{kahneman73attention, oviatt03book}.
This allows annotators to concurrently solve multiple tasks, such as providing the semantic label and the location of an object.
In fact, people naturally choose to point for providing spatial information and to speak for semantic information when using multimodal interfaces \cite{oviatt03book}.
(v) it makes the interface design extremely simple, which is what allows to combine \taskname{} and bounding box annotation into a single task.

Using speech as an input modality, however, poses certain challenges.
In order to extract object annotations from speech and mouse inputs, several technical challenges need to be tackled.
These include transcribing the speech, inferring class labels, and aligning them with object location annotations (Sec.~\ref{sec:postprocessing}).
Furthermore, as speech is free-form in nature, annotators need to be trained to know the class vocabulary to be annotated,
in order to not label other objects or forget to annotate some classes.
In Sec.~\ref{sec:training} \& \ref{sec:joint_training} we address these challenges, which allows us to design annotation interfaces for fast and accurate labelling.

We validate our approach with extensive experiments (Sec. \ref{sec:experiments} \& \ref{sec:experiments_xclick}).
In particular we:

\begin{itemize}%

  \item Show that speech enables fast \taskname{}: \att{2.3${\times}$} faster on the \coco{} dataset \cite{lin14eccv} than the hierarchical approach of \cite{lin14eccv}, and \att{14.9$\times$} faster than \cite{deng14chi} on \ilsvrc{} \cite{russakovsky15ijcv}.
  
  \item Show that the class labelling can be embedded into the bounding box annotation stage, which allows to produce class labels at zero additional cost. On COCO, this makes the overall process \att{$1.9\times$} faster than the two-stage approach.
  
  \item Demonstrate that our method scales to large vocabularies.
  
  \item Show that through our training task annotators learn to use the provided vocabulary for naming objects with high fidelity.
  
  \item Analyze the accuracy of models for automatic speech recognition (ASR) and show that it supports deriving high-quality annotations from speech.
  
\end{itemize}

This paper is an extension of our preliminary work \cite{gygli19cvpr}, which focused only on object class labelling.
It introduces a new annotation protocol to simultaneously annotate objects with bounding boxes and class labels (Sec.~\ref{sec:xclick_task}), new experiments (Sec.~\ref{sec:experiments_xclick}),
and a better method for temporally segmenting and aligning speech with object location annotations (Sec.~\ref{sec:postprocessing}).

\section{Related Work}
Using speech as an input modality has a long history \cite{bolt80siggraph} and is recently emerging as a research direction in Computer Vision\cam{ \cite{dai16thesis, vasudevan17cvpr, vaidyanathan18acl, harwath18eccv}}.
To the best of our knowledge, however, our paper is the first to show that speech allows for more efficient \taskname{} than \cite{lin14eccv,deng14chi} and enables simultaneous \jointtask{}.
We now discuss previous works in the areas of leveraging speech, efficient \taskname{}, learning from point supervision
and bounding box annotation.

\para{Leveraging speech inputs.}
To point and speak is an efficient and natural way of human communication.
Hence, this approach was quickly adopted when designing computer interfaces: as early as 1980, \cite{bolt80siggraph} investigates using speech and gestures for manipulating shapes.
Most previous works in this space analyse what users choose when offered different input modalities \cite{hauptmann89sigchi,oviatt96sigchi,oviatt97integration,oviatt03book},
while only few approaches focus on the added efficiency of using speech.
The most notable such work is \cite{pausch91vio}, which measures the time needed to create a drawing in MacDraw.
They compare using the tool as is, which involves selecting commands via the menu hierarchy, to using voice commands. They show that using speech gives an average speedup of 21\% and mention this is a ``lower bound'', as the tool was not designed with speech in mind.

In Computer Vision, \cite{vasudevan17cvpr} detect objects given spoken referring expressions,
while \cite{harwath18eccv} learn an embedding from spoken image-caption pairs. Their approach obtains promising first results, but still performs inferior to learning on top of textual captions produced by Google's automatic speech recognition.
 \cite{damen18eccv} annotates the EPIC-KITCHENS dataset based on spoken free-form narratives, which cover some of the objects present in the image.
These narratives are however transcribed {\em manually}, and then object class labels are derived from transcribed nouns, again manually.
Instead, our approach is fully automatic and we exhaustively label all objects from a given vocabulary.
Finally, more closely related to our work, \cite{vaidyanathan18acl} re-annotated a subset of \coco{} with spoken scene descriptions and human gaze.
While efficient, free-form scene descriptions are noisier when used for \taskname{}, as annotators might refer to objects with ambiguous names, mention nouns that do not correspond to objects shown in the image \cite{vaidyanathan18acl}, or there might be inconsistencies in naming the same object classes across different annotators.
Our approach avoids the additional complexities of parsing free-form sentences to extract object names and gaze data to extract object locations.

\para{Efficient \taskname{}.}
The na\"ive approach to annotating the presence of object classes grows linearly with the
size of the vocabulary (one binary present / absent question per class).
The idea behind sub-linear schemes is to group the classes into meaningful super-classes, such that several of them can be ruled out at once. If a super-class (\eg animals) is not present in the image, then one can skip the questions for all its subclasses (cat, dog, \textit{etc.}).
This grouping of classes can have multiple levels.
The annotation schemes behind \coco{} \cite{lin14eccv} and \ilsvrc{} \cite{deng14chi,russakovsky15ijcv} datasets both fall into this category,
but they differ in how they define and use the hierarchy.

\ilsvrc{} \cite{russakovsky15ijcv} was annotated using a series of hierarchical questions \cite{deng14chi}.
For each image, 17 top-level questions were asked (\eg ``Is there a living organism?''). For groups that are present, more specific questions are asked subsequently, such as ``Is there a mammal?'', ``Is there a dog?'', \etc
The sequence of questions for an image is chosen dynamically, such that the they allow to eliminate the maximal number of labels at each step \cite{deng14chi}.
This approach, however, involves repeated visual search, in contrast to ours, which is guided by the annotator scanning the image for objects, done only once.
Overall, this scheme takes close to 3 minutes per image \cite{russakovsky15cvpr} for annotating the 200 classes of \ilsvrc{}.
On top of that, constructing such a hierarchy is not trivial and influences the final results \cite{russakovsky15ijcv}.

In the protocol used to create \coco{} \cite{lin14eccv}, annotators are asked to mark one object for each class present in an image
by choosing its symbol from a two-level hierarchy and dragging it onto the object (Fig. \ref{fig:hierarchical_interface}).
While this allows to take the image, rather than the questions as the root of the labelling task, it requires repeatedly searching for the right class in the hierarchy, which induces significant time cost.
In our interface, such an explicit class search is not needed, which speeds up the annotation process.

Rather than using a hierarchy, \cite{kuznetsova18arxiv} annotated object class labels on the Open Images dataset by relying on an image classifier.
The classifier creates a shortlist of object classes likely to be present, which are then verified by annotators using binary questions.
The shortlist is generated using a pre-defined threshold on the classifier scores. Thus, this approach trades off completeness for speed.
In practice, \cite{kuznetsova18arxiv} asks annotators to verify 10 out of 600 classes, but report a rather low recall of 59\%, despite disregarding ``difficult'' objects in evaluation.

\para{Point supervision.}
The output of our interface for \taskname{} is a list of all classes present in the image with a point on one object for each.
This kind of labelling is efficient and provides useful supervision for several image \cite{papadopoulos17cvpr, bearman16eccv, laradji18arxiv} and video \cite{mettes16eccv,manen17iccv} object localization tasks. %
In particular, \cite{papadopoulos17cvpr, bearman16eccv, manen17iccv} show that for their task, point clicks deliver better models than other alternatives when given the same annotation budget.

\para{Bounding box annotation.}
Typically, bounding boxes are annotated given image-level labels \cite{lin14eccv,su12aaai, russakovsky15ijcv}.
\cite{su12aaai} reports that it takes 25.5 seconds to draw a bounding box.
Recently, \cite{papadopoulos17iccv} proposed extreme clicking, where a bounding box is annotated by clicking on four extreme points of the object.
Using this procedure makes bounding box annotation significantly faster: it takes only 7.4 seconds on average to draw bounding boxes around for the objects in the Open Images dataset \cite{kuznetsova18arxiv}.
Hence, we also use extreme clicking in our simultaneous \jointtask{}.

\begin{figure}[t]
		\centering
		\includegraphics[width=1.0\linewidth]{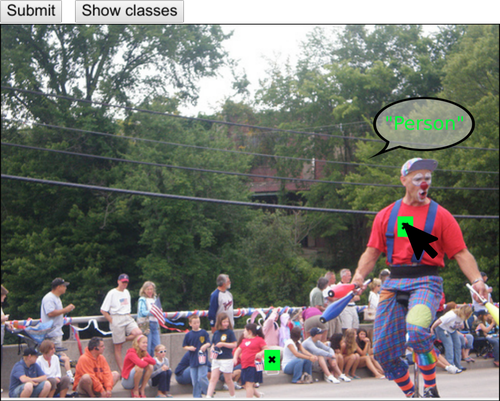}
		\caption{\textbf{Our interface for \taskname{}.} Given an image the annotator is asked to click on one object per class and say its name.
		To aid memory, we additionally allow to review the class vocabulary through the ``Show classes'' button.
		}
	\label{fig:interface}
\end{figure}

\section{Object Class Labelling}
\label{sec:object_class_labelling}

We now describe our interface for using speech in the first annotation stage: determining the presence or absence of object classes in an image (Sec.~\ref{sec:main_task}).
Before annotators can proceed to the main task, we require them to pass a training stage. This helps them memorise the class vocabulary and get confident with using the interface (Sec.~\ref{sec:training}).
From the output of this annotation task we derive class labels by transcribing the recorded speech and mapping it to class names as described in Sec.~\ref{sec:speech_to_class}.

\subsection{Annotation task}
\label{sec:main_task}

First, annotators are presented with the class vocabulary and instructed to memorise it.
Then, they are asked to label images with object classes from the vocabulary, by scanning the image and saying the names of the different classes they see.
Hence, this is a simple visual search task that does not require any context switching.
While we are primarily interested in object class labels, we ask annotators to click on one object for each class, as the task naturally involves finding objects anyway.
This matches the COCO protocol, allowing for direct comparisons (Sec.~\ref{sec:experiments_baseline}).
It further provides information that can be used as input to weakly-supervised methods \cite{bearman16eccv,papadopoulos17cvpr}.
Fig.~\ref{fig:interface} shows the interface with an example image.

To help annotators restrict the labels they provide to the predefined vocabulary, we allow them to review it using a button that shows all class names including their symbols.
\begin{figure}[t]
\centering
   \begin{subfigure}[b]{1\linewidth}
   \centering
		\includegraphics[width=1.0\linewidth]{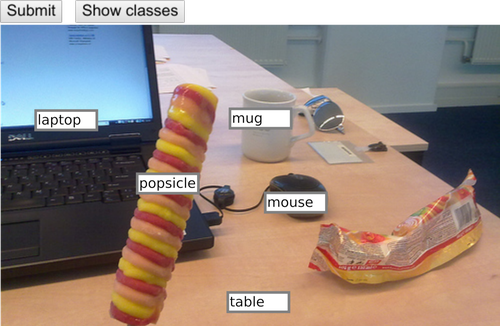}
        \caption{}
        \label{fig:qualification_task}
        \vspace{0.2cm}
    \end{subfigure}
   \begin{subfigure}[b]{1\linewidth}
		\includegraphics[width=1\linewidth]{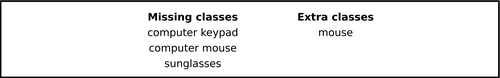}		
        \caption{}
        \label{fig:qualification_feedback}
    \end{subfigure}
		\caption{\textbf{Training process for \taskname{}}. \ref{fig:qualification_task} shows the training task: marking an object per class with a click and saying and writing its name.
		\ref{fig:qualification_feedback} shows the feedback given after each image.
		}
	\label{fig:qualification}
\end{figure}

\subsection{Annotator training}
\label{sec:training}
Before tackling the main task, annotators go through a training stage which gives feedback after every image and also aggregated statistics after 80 images.
If they meet our accuracy targets, they can proceed to the main task. If they fail, they can repeat the training until they succeed.

\para{Purpose of training.}
\cam{Training helps annotators to get confident with an interface and allows to ensure they correctly solve the task and produce high-quality labels.
As a consequence, it has become common practice \cite{russakovsky15ijcv,su12aaai, lin14eccv, kuznetsova18arxiv, papadopoulos17cvpr}.%
}

While we want to annotate classes from a predefined vocabulary, speech is naturally free-form.
In our initial experiments we found that annotators produced lower recall compared to an interface which displays an explicit list of classes, due to this discrepancy.
Hence, we designed our training task to ensure annotators memorise the vocabulary and use the correct object names.
Indeed, after training annotators with this process they rarely use object names that are not in the vocabulary and obtain a high recall, comparable to \cite{lin14eccv} (Sec.~\ref{sec:experiments_coco} \& ~\ref{sec:detailed_experiments}).

\para{Training procedure.}
The training task is similar to the main task, but we additionally require annotators to type the words they say (Fig.~\ref{fig:qualification_task}).
This allows to measure transcription accuracy and dissect different sources of error in the final class labelling (Sec.~\ref{sec:detailed_experiments}).
After each image we give immediate feedback listing their mistakes, by comparing their answers against a pre-annotated ground truth.
This helps annotators memorise the class vocabulary and learn to spot all object classes (Fig.~\ref{fig:qualification_feedback}).
We base this feedback on the written words, rather than the transcribed audio, for technical simplicity.

\para{Passing requirements.}
At the beginning of training, annotators are given targets on the minimum recall and precision they need to reach.
Annotators are required to label 80 images and are given feedback after every image, listing their errors on that image, and on how well they do overall with respect to the given targets.
If they meet the targets after labelling 80 images, they successfully pass training.
In case of failure, they are allowed to repeat the training as many times as they want.

\section{Simultaneous Bounding Box and Class Labelling}
\label{sec:xclick_task}

 We propose an interface to simultaneously annotate bounding boxes and class labels,
 thus combining the two standard stages into one.
 Before annotators can proceed to the main task, we require them to pass a training stage (Sec.~\ref{sec:joint_training}).
 The annotation task produces a series of bounding boxes with start and end times and an audio recording for each image.
 We will transform this data into the final object annotations by deriving object classes from speech and matching them to bounding boxes as described in Sec.~\ref{sec:postprocessing}.

  \subsection{Annotation task}
  \label{sec:joint_task}
  
\begin{figure}[t]
    \centering
    \includegraphics[width=1.0\linewidth]{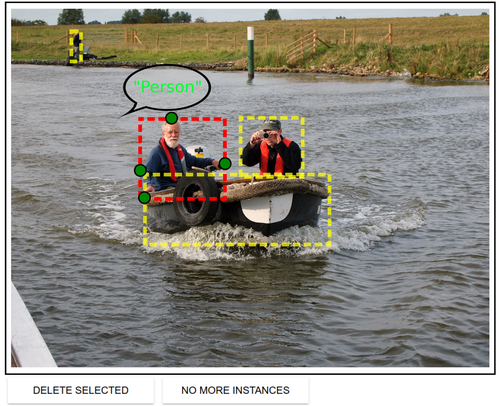}
    \caption{\textbf{Our interface for simultaneous \jointtask{}.} Annotators are asked to mark all objects of a given vocabulary with a bounding box and their class names.
    }
    \label{fig:interface_xclick}
  \end{figure}

  As in Sec.~\ref{sec:main_task}, annotators are presented with the class vocabulary and instructed to memorise it (but can again review it later).
  Then, they are asked to annotate
  \emph{all} objects of all classes that are in the vocabulary.
  For each object, the annotator simultaneously draws a bounding box while saying its class name.
  We annotate bounding boxes using the efficient extreme clicking method~\cite{papadopoulos17iccv}, which requires clicking on the top, bottom, left- and right-most point of an object.
  Hence, this annotation task requires speaking and clicking, which can naturally be done in parallel \cite{kahneman73attention, oviatt03book}.
  Fig.~\ref{fig:interface_xclick} shows the interface with example annotations.

  \subsection{Annotator training}
  \label{sec:joint_training}
  We train annotators for the task of drawing bounding boxes and saying object class names.
  Thereby we use two training steps. The first trains annotators to quickly draw accurate bounding boxes of a given object class, following the training protocol of~\cite{papadopoulos17iccv,kuznetsova18arxiv}.
  The second step trains annotators to say the correct class names.
  It is similar to the training procedure for \taskname{} (Sec.~\ref{sec:training}),
  except that annotators have to mark an objects location with a bounding box, rather than just a single click.

\section{Transcription and temporal alignment}
\label{sec:postprocessing}

  \begin{figure}[t]
    \centering
    \includegraphics[width=0.8\linewidth]{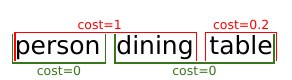}
    \caption{\textbf{Transcription segmentation.} Our method assigns a cost to each subsequence of words.
    Then, it segments the transcription into the list of object class names that has minimal cost.
    In this example, segmenting into $\{\mathrm{person}, \mathrm{dining\ table}\}$ has zero cost, while all other segmentations
    have cost $ > 0$, as other possible subsequences such as ``person dining'' are not in the vocabulary.
    }
    \label{fig:segmentation_to_classnames}
  \end{figure}

  The interfaces presented in Sec.~\ref{sec:object_class_labelling} \& Sec.~\ref{sec:xclick_task} both output an audio recording and a sequence of object locations $(o_1,o_2,\dots)$, each with a time interval. The object location is a single click in Sec.~\ref{sec:object_class_labelling} and a bounding box in Sec.~\ref{sec:xclick_task}.
  To produce the final object annotations, we first transcribe and segment the audio to derive a sequence of object class labels (Sec.~\ref{sec:speech_to_class}).
  Then, we temporally align the class labels to object locations using global sequence alignment (Sec.~\ref{sec:alignment}).
  The same alignment algorithm applies generally to both Sec.~\ref{sec:object_class_labelling} \&~\ref{sec:xclick_task}.

  \subsection{Infer class names from speech}
  \label{sec:speech_to_class}
  
  We transcribe the audio with Google's automatic speech recognition API\footnote{\url{https://cloud.google.com/speech-to-text/}}.
  The API outputs a sequence of utterances (continuous speech blocks, enclosed by pauses) with a ranked list of transcription alternatives for each.
  We first segment the different transcriptions into class names and then choose the most likely transcription for each utterance.

  \para{Segmenting a transcription.}
  While the transcription for an utterance would ideally only contain a single class name, we find that in practice it sometimes contains two or more classes, \eg ``person dining table''.
  This happens when the pause between saying different class names is too short.
  Thus, we propose an algorithm to automatically segment the transcription into the most likely sequence of class names.

  \changedtext{
We assign a cost to each subsequence of words,~\eg ``person'', ``person dining'', \etc (Fig.~\ref{fig:segmentation_to_classnames}).
  The cost corresponds to the dissimilarity between the subsequence and the nearest class name in the vocabulary.
  Thereby we represent a subsequence with its word2vec \cite{mikolov13arxiv} vector and use the
  cosine distance between word vectors as a dissimilarity measure.
  For subsequences that are in the vocabulary,~\eg ``person'' and ``dining table'', the cost is thus zero.
  For subsequences that are semantically similar to words in the vocabulary,~\eg ``table'' is similar to ``dining table'', the cost is above zero, but still low (0.2 in the figure).
  For subsequences that do not even
  correspond to a valid noun,~\eg ``person dining'', we assign a fixed high cost (1 in the figure). 

 The total cost $\epsilon(s)$ of a segmentation $s$ is the sum over the cost of each subsequence in~$s$.    
 By relying on word2vec distance, we assign low costs to synonyms of vocabulary words.~\eg a segmentation of ``person sofa'' into $\{\mathrm{person}, \mathrm{sofa}\}$ would correctly get a low total cost even when the vocabulary contains ``couch'' instead of ``sofa''.

  We find the most likely segmentation $s^*$ of a transcription as the one with the lowest total cost using dynamic programming.
  In the example from Fig.~\ref{fig:segmentation_to_classnames}, this would be $s^*=\{\mathrm{person}, \mathrm{dining\ table}\}$, which has segmentation cost $\epsilon(s^*)=0$.
} 

  \para{Selecting the most likely transcription.}
  An utterance comes with multiple possible transcriptions, \eg ``person dining table'',``ocean dining table'', ``person dying table''.
  We re-rank these transcriptions by their minimum segmentation cost (see above) and choose the one with the lowest cost.
  If two transcriptions have the same cost, we choose the one which the speech recognition API ranked higher.
  In the above example, this would allow to correctly identify ``person dining table'' as the correct transcription,
  as its most likely segmentation $\{\mathrm{person}, \mathrm{dining\ table}\}$ contains only class names from the vocabulary (and thus has zero cost).
  The other alternatives have a non-zero cost, as ``ocean'' and ``dying table'' are not part of the vocabulary.
  
  In rare cases the top transcription contains an object name that is not in the vocabulary.
  In these cases we map the object name to the closest class in the vocabulary, using the cosine distance of their word2vec \cite{mikolov13arxiv} representation.

  The final output of this algorithm is a sequence of class labels $(c_1, c_2, \dots)$,
  each with an associated start and end time (we can do this as the speech recognition API outputs the start and end time of each word).

  \subsection{Aligning class labels and object locations}
  \label{sec:alignment}
  \begin{figure}[t]
    \centering
    \includegraphics[width=1\linewidth]{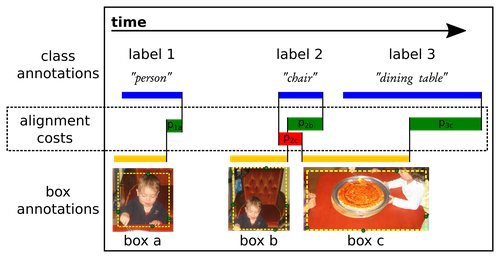}
    \caption{\textbf{Temporal alignment.}
    Object locations and class labels are aligned using the time intervals during which they are annotated.
    The cost of an alignment depends on the temporal overlap of saying the class name and providing the object location.
    We find a globally optimal alignment with the Needleman-Wunsch algorithm \cite{needleman70jmb}.
    In above example, the correct alignment of label 2 with box b has higher cost than aligning it with box c (when considering this on its own).
    Using a global alignment technique allows to correctly align them nonetheless.}
    \label{fig:temporal_alignment}
  \end{figure}
  We propose a method for temporally aligning the sequence of class labels $(c_1, c_2, \dots)$
  with object locations $(o_1,o_2,\dots)$.
  As annotators sometimes speak before or after annotating the object location, rather than during it,  aligning the two is not trivial.
  
  Hence, we align the elements in the two sequences based on how much they temporally overlap (Fig.~\ref{fig:temporal_alignment}). %
  The cost of an alignment is the sum of aligning individual elements, plus a gap penalty for each element that has no correspondence.
  Gaps can happen in practice,~\eg if an annotator discards a bounding box annotation,  which would lead to a class label with no correspondence.
  We define the cost of aligning two elements as $\rho(c_i, o_j) = 1 - \mathrm{i}(c_i, o_j) / \mathrm{d}(c_i)$,
  where $\mathrm{i}(c_i,o_j)$ is the temporal overlap of the object location and class label annotation times.
  $\mathrm{d}(c_i)$ is the duration of saying class label $c_i$.
  The cost thus encourages aligning elements where the class name was said during the time interval at which the object location was drawn. But it does so smoothly, allowing for deviations from the ideal case.

  We find the optimal global alignment by relying on the Needleman-Wunsch algorithm \cite{needleman70jmb}.
  Thereby each class label $c_i$ and object location $o_j$ can be matched at most once and gaps are possible.
  The algorithm uses dynamic programming to find the best global alignment, which is the one that has the minimum cost.
  We found a large range of gap penalties to work well and empirically set it to \att{$0.5$}.

  This alignment algorithm outputs a sequence of object annotations, each consisting of a class label $c$ and an object location $o$.

\section{Experiments on Object Class Labelling}
\label{sec:experiments}

Here we present experiments on annotating object class labels using our speech-based interface and the hierarchical interface of \cite{lin14eccv}.
First, in Sec.~\ref{sec:experiments_baseline} we re-implement the interface of \cite{lin14eccv} and compare it to the official reported results in \cite{lin14eccv}.
Then, we compare the two interfaces on the COCO dataset, where the vocabulary has 80 classes (Sec.~\ref{sec:experiments_coco}). In Sec.~\ref{sec:experiments_imagenet} we scale up annotation to a vocabulary of 200 classes by experimenting on the \ilsvrc{} dataset.
Finally, Sec.~\ref{sec:detailed_experiments} provides additional analysis such as the transcription and click accuracy as well as response times per object.

\subsection{Hierarchical interface of \cite{lin14eccv}}

\label{sec:experiments_baseline}
\begin{figure}[t]
	\centering\includegraphics[width=1\linewidth]{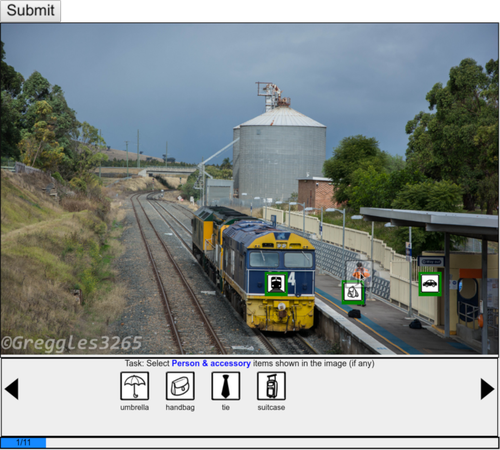}
		\caption{Our re-implementation of the hierarchical interface of \cite{lin14eccv} for \taskname{}.}
	\label{fig:hierarchical_interface}
\end{figure}
In the interface of \cite{lin14eccv}, annotators mark one object for each class present in an image by choosing its symbol from a two-level hierarchy and dragging it onto the object. While \cite{lin14eccv} reports coarse timings, we opted to re-implement their interface for fair comparison and to do a detailed analysis on how annotation time is spent (Fig.~\ref{fig:hierarchical_interface}).
First, we made five crowd workers pass a training task equivalent to that used for our interface (Sec.~\ref{sec:training}).
Then, they annotated a random subset of 300 images of the \coco{} validation set (each image was annotated by all workers).

\para{Results.}
Annotators take \att{$29.9$} seconds per image on average, well in line with the $27.4$ seconds reported in \cite{lin14eccv}.
Hence, we can conclude that our implementation is equivalent in terms of efficiency.

Annotators have produced labels with \att{89.3\%} precision and \att{84.7\%} recall against the ground truth (Tab.~\ref{tab:accuracy}).
Thus, they are accurate in the labels they produce and recover most object classes. We also note that the \coco{} ground truth itself is not free of errors, hence limiting the maximal achievable performance.
Indeed, our recall and precision are comparable to the numbers reported in \cite{lin14eccv}.

\para{Time allocation.}
In order to better understand how annotation time is spent, we recorded mouse and keyboard events.
This allows us to estimate the time spent on searching for the right object class in the hierarchy of symbols and measure the time spent dragging the symbol.
On average, search time is \att{$14.8$}s and drag time \att{$3.4$}s per image.
Combined, these two amount to \att{$61\%$} of the total annotation time, while the rest is spent on other tasks such as visual search.
This provides a target on the time that can be saved by avoiding these two operations, as done in our interface. In the remainder of this section, we compare our speech-based approach against this annotation method.

\begin{figure}[t]
	\centering\includegraphics[width=1\linewidth]{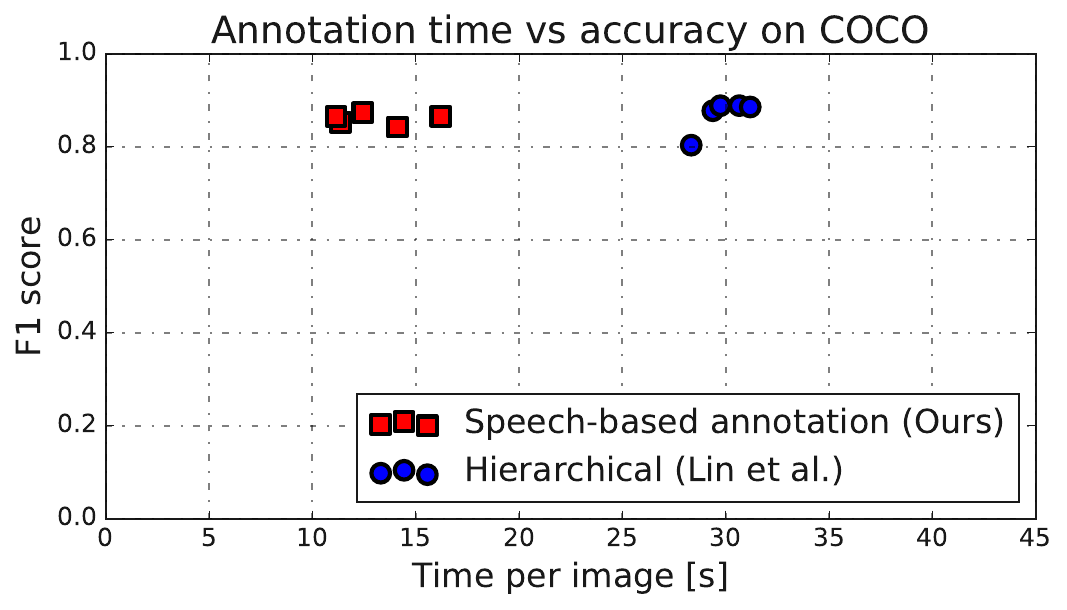}
		\vspace{-0.2cm}
		\caption{Our approach \vs the hierarchical interface of \cite{lin14eccv}.
		Each point in the plot corresponds to an individual annotator. F1 score is the harmonic mean between recall and precision. Dataset: COCO.}
	\label{fig:results_coco}
\end{figure}

\subsection{Our interface on COCO}
\label{sec:experiments_coco}

We evaluate our approach and compare it to \cite{lin14eccv}.
Annotations with our interface were done by a new set of crowd workers,
to avoid bias arising from having used the hierarchical interface before.
The workers are all Indian nationals and speak English with an Indian accent.
Hence, we use a model of Indian English for the automatic speech recognition.
We also provide the class vocabulary as phrase hints\footnote{\url{https://cloud.google.com/speech-to-text/docs/basics\#phrase-hints}},
which is crucial for achieving high transcription accuracy (Sec.~\ref{sec:detailed_experiments}).

\para{Speed and semantic accuracy.}
Fig.~\ref{fig:results_coco} and Tab.~\ref{tab:accuracy} show results. Our method achieves a speed-up of \att{$2.3\times$} over \cite{lin14eccv} at similar F1 scores (harmonic mean of precision and recall).
In Sec.~\ref{sec:experiments_baseline} we estimated that annotation could be sped up by up to \att{$2.6\times$} by avoiding symbol search and dragging.
Interestingly, our interface achieves a speedup close to this target, confirming its high efficiency.

Despite the additional challenges of handling speech, average precision is only \att{1.7\%} lower than for \cite{lin14eccv}.
Hence, automatic speech transcription does not affect label quality much (we study this further in Sec.~\ref{sec:detailed_experiments}).
Recall is almost identical (\att{0.5\%} lower), confirming that, thanks our training task (Sec.~\ref{sec:training}), annotators remember what classes they have to label.

\para{Location accuracy.}
We further evaluate the location accuracy of the clicks by using the ground-truth segmentation masks of~\coco{}. Specifically, given an object annotation with class $c_i$, we evaluate whether its click position lies on a ground-truth segment of class $c_i$.
If class $c_i$ is not present in the image at all, we ignore that click in the evaluation to avoid confounding semantic and location errors.

This analysis shows that our interface leads to high location accuracy: \att{$96.1\%$} of the clicks lie on the object.
For the hierarchical interface it is considerably lower at \att{$90.7\%$}.
While this may seem surprising, it can be explained by the differences in the way the location is marked.
In our interface one directly \textit{clicks} on the object, while \cite{lin14eccv} requires \textit{dragging} a relatively large, semi-transparent class symbol onto it (Fig.~\ref{fig:hierarchical_interface}).

Parts of the speed gains of our interface are due to concurrently providing semantic and location information.
However, this could potentially have a negative effect on click accuracy.
To test this, we compare to the click accuracy that the annotators in \cite{bearman16eccv} obtained on the PASCAL VOC dataset.
Their clicks have a location accuracy of \att{96.7\%}
comparable to our \att{$96.1$\%}, despite the simpler dataset with larger objects on average, compared to COCO.
Hence, we can conclude that clicking while speaking does not negatively affect location accuracy.

\subsection{Our interface on \ilsvrc{} 2014}
\label{sec:experiments_imagenet}

Here we apply our interface and the hierarchical interface of \cite{lin14eccv} to a larger vocabulary of 200 classes, using 300 images from the validation set of \ilsvrc{} \cite{russakovsky15ijcv}.
For \cite{lin14eccv} we manually constructed a two-level hierarchy of symbols, based on the multiple hierarchies supplied by \cite{russakovsky15ijcv}.
The hierarchy consists of 23 top-level classes, such as ``fruit'' and ``furniture'', each containing between 5 to 16 object classes.
\begin{table}[t]
	\centering
	\resizebox{1\linewidth}{!}{%
	\setlength{\tabcolsep}{2pt}
	\begin{tabular}{|c|r|r|r|}
		\hline
		&  \textbf{Speech} & \textbf{\cite{lin14eccv}} & \textbf{\parbox{1.7cm}{\cite{deng14chi}}} \\
		\hline
		\multicolumn{4}{|c|}{\textbf{\coco{}}} \\
		Recall     & 84.2 \% & 84.7 \% & \\
		Precision  & 87.6 \% & 89.3 \% & \\
		\hdashline
		Time / image & 13.1s & 29.9s & \\
		Time / label & 4.5s & 11.5s & \\
		\hline
		\multicolumn{4}{|c|}{\textbf{\ilsvrc{}}} \\
		Recall     & 83.2 \% & 88.6 \% & \\
		Precision  & 80.3 \% & 76.6 \% & \\
		\hdashline
		Time / image & 12.0 sec. & 31.1 sec. & $\approx$ 179 sec.\\
		Time / label & 7.5 sec. & 18.4 sec. & $\approx$ 110 sec.\\
		\hline
	\end{tabular}
	}
	\caption{Accuracy and speed of our interface (Speech) and hierarchical approaches \cite{lin14eccv, deng14chi} for \taskname{}.
	Our interface is significantly faster at comparable label quality. Timings for \cite{deng14chi} are taken from \cite{russakovsky15cvpr}.
	Note that the numbers for Speech differ slightly from those reported in \cite{gygli19cvpr}, due to the changes
	in the temporal segmentation and alignment (Sec.~\ref{sec:postprocessing}).
	}
	\label{tab:accuracy}
\end{table}
\begin{figure}[t]
	\centering\includegraphics[width=1\linewidth]{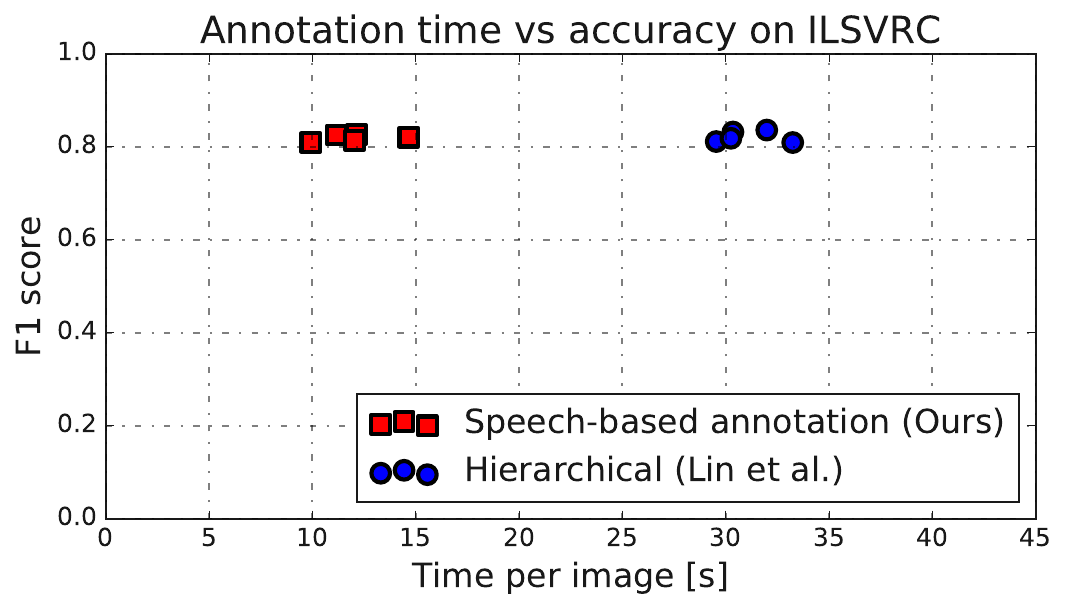}
	\vspace{-0.2cm}
	\caption{Our approach \vs the hierarchical interface \cite{lin14eccv}. Each point in the plot corresponds to an individual annotator. Dataset: \ilsvrc{}.}
	\label{fig:results_imagenet}
\end{figure}

\begin{figure*}[t]
	\centering\includegraphics[width=0.25\linewidth,trim={0 0 0 0},clip]{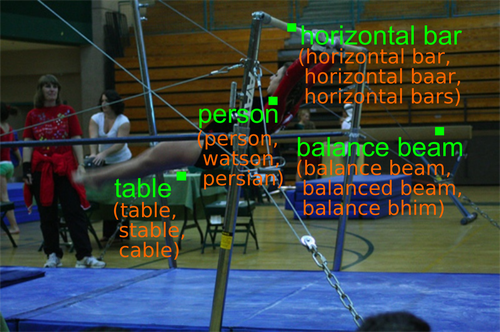}%
	\centering\includegraphics[width=0.25\linewidth,trim={0 0 0 0},clip]{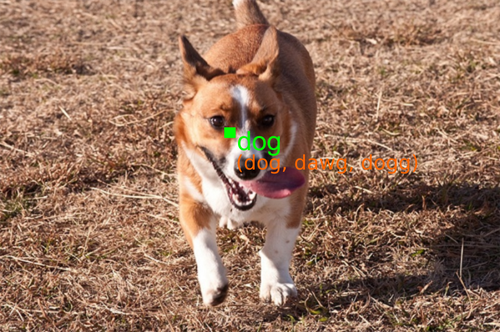}%
	\centering\includegraphics[width=0.25\linewidth,trim={0 1.9cm 0 0},clip]{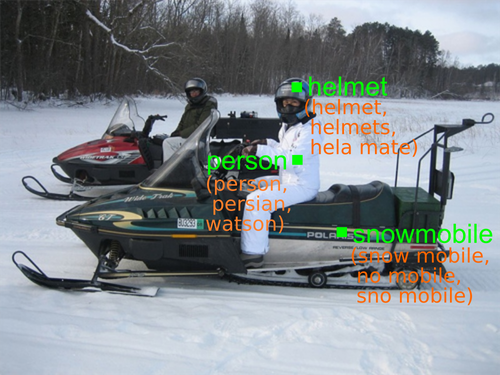}%
	\centering\includegraphics[width=0.25\linewidth,trim={0 0.95cm 0 0},clip]{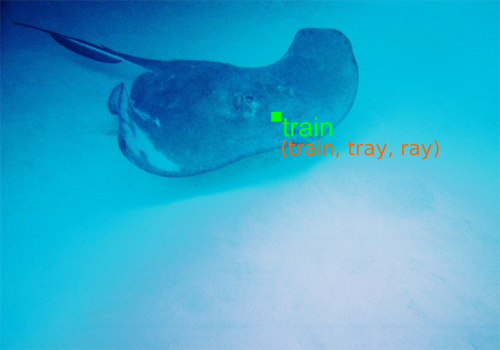}%
	\caption{Example annotations on \ilsvrc{}.
	For each click we show the three alternatives from the ASR model (\textcolor{orange}{orange}) and the final class label (\textcolor{green}{green}).
	The first three images show typical annotations produced by our method.
	The last one shows a failure case:
	while the correct name is among the alternatives, an incorrect transcription matching a class name ranks higher, hence the final class label is wrong.
	}

	\label{fig:examples}
\end{figure*}

\para{Speed and semantic accuracy.}
Fig.~\ref{fig:results_imagenet} shows a comparison to \cite{lin14eccv} in terms of speed and accuracy, while 
Fig.~\ref{fig:examples} shows example annotations produced with our interface.
In Tab.~\ref{tab:accuracy}, we also compare to the speed of \cite{deng14chi}, the method that was used to annotate this dataset. Our approach is substantially faster than both: \att{2.6$\times$} faster than \cite{lin14eccv} and \att{14.9$\times$} faster than \cite{deng14chi}.
We also note that \cite{deng14chi} only produces a list of classes present in an image, while our interface and \cite{lin14eccv} additionally yield the location of one object per class.

Despite the increased difficulty of annotating this dataset, which has considerably more classes than~\coco{},
annotators produce high-quality labels with our interface. The F1 score is similar to that of \cite{lin14eccv} (\att{81.7\%} \vs \att{82.2\%}). While recall is lower for our interface, precision is higher.

\begin{figure}[t]
	\centering\includegraphics[width=1\linewidth]{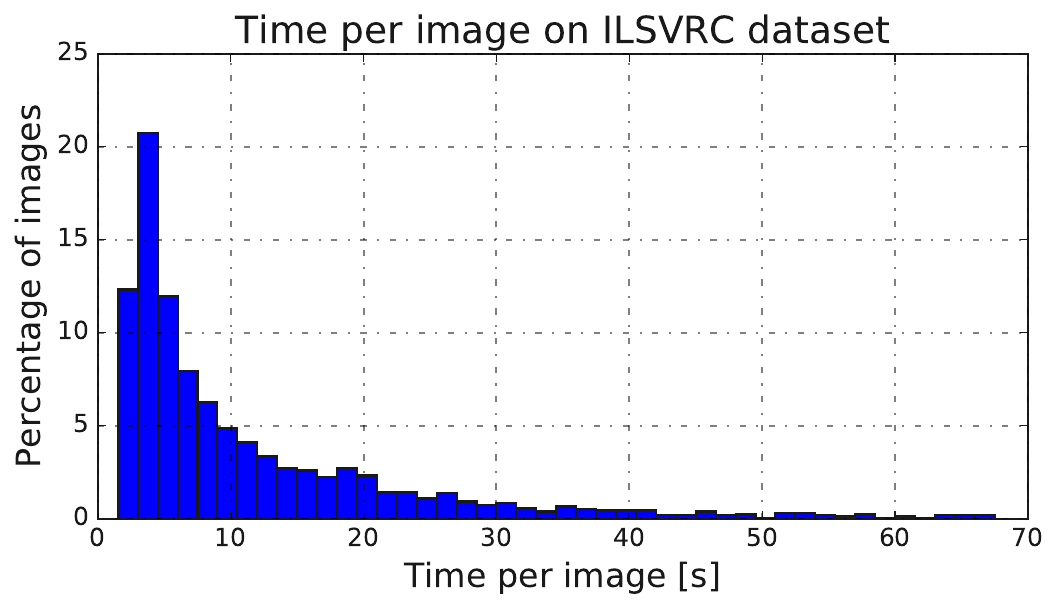}
	\vspace{-0.2cm}
	\caption{Histogram of the time required to annotate an image using our interface. Dataset:~\ilsvrc{}.}
	\label{fig:time_per_image_imagenet}
\end{figure}

Fig.~\ref{fig:time_per_image_imagenet} shows a histogram of the annotation time per image. Most images are annotated extremely fast, despite the large vocabulary, as most images in this dataset contain few classes.
Indeed, there is a strong correlation between the number of object classes present in an image and its annotation time (rank correlation \att{0.55}).
This highlights the advantage of methods that are rooted on the image content, rather than the vocabulary: their annotation time is low for images with few classes. Instead, methods rooted on the vocabulary cannot exploit this class sparsity to a full extent.
The na\"ive approach of asking one yes/no question per class is actually even slower the fewer objects are present, as determining the absence of a class is slower than confirming its presence \cite{ehinger09modelling}.

\subsection{Additional analysis of our interface}
\label{sec:detailed_experiments}

\para{Time allocation.}
To understand how much of the annotation time is spent on what, we analyse timings for speaking and moving the mouse on the~\ilsvrc{} dataset.
Of the total annotation time, \att{26.7\%} is spent on speaking.
The mouse is moving \att{74.0\%} of the total annotation time, and \att{62.4\%} of the time during speaking.
The rather high percentage of time the mouse moves during speaking confirms that humans can naturally carry out visual processing and speaking concurrently. %

In order to help annotators label the correct classes, we allowed them to consult the class vocabulary, through a button on the interface (Fig.~\ref{fig:interface}).
This takes \att{7.2\%} of the total annotation time, a rather small share.
Annotators consult the vocabulary in fewer than \att{20\%} of the images. %
When they consulted it, they spent \att{7.8} seconds looking at it, on average.
Overall, this shows the annotators feel confident about the class vocabulary and confirms that our annotator training stage is effective.

\begin{figure}[t]
	\centering\includegraphics[width=1\linewidth]{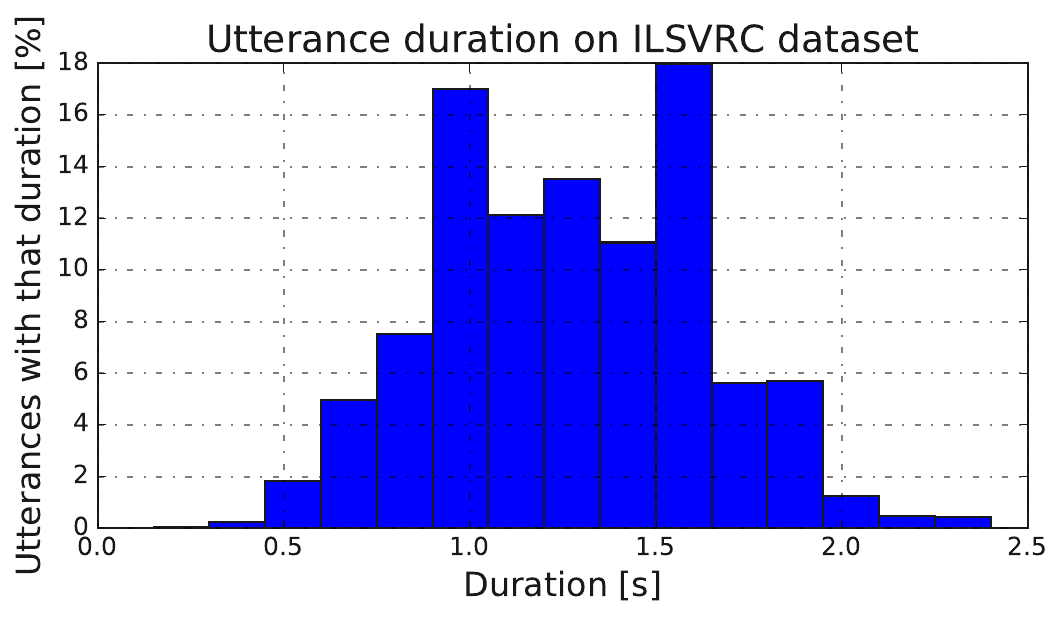}
	\vspace{-0.2cm}
	\caption{Histogram of the time spent \textit{saying} the object name on \ilsvrc{}.
	Saying the object names is fast and usually takes less than 2 seconds.
	}
	\label{fig:utterance_durations}
\end{figure}

In addition, we analyse the time it takes annotators to say an object name in Fig.~\ref{fig:utterance_durations}, which shows a histogram of speech durations. As can be seen, most names are spoken in \att{$0.5$ to $2$ seconds}.

\begin{figure}[t]
	\centering\includegraphics[width=1\linewidth]{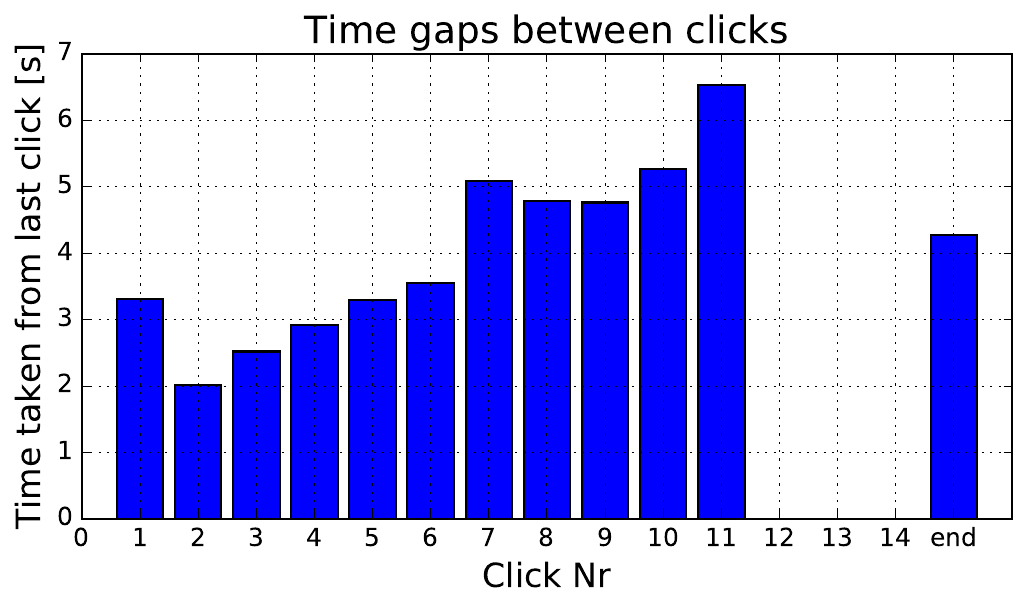}
	\vspace{-0.2cm}
	\caption{Analysis of the time it takes for the first and subsequent clicks when annotating object classes on the \coco{} dataset.}
	\label{fig:time_per_click}
\end{figure}

\para{Per-click response time.}
In Fig.~\ref{fig:time_per_click} we analyse the time taken to annotate the first and subsequent classes of an image in the \coco{} dataset.
It takes \att{3.3}s to make the first click on an object, while the second takes \att{2.0}s only.
This effect was also observed by \cite{bearman16eccv}.
Clicking on the first object incurs the cost of the initial visual search across the whole scene, while the second is a continuation of this search and thus cheaper \cite{watson07eye, rayner09, lleras05rapid}.
After the second class, finding more classes becomes increasingly time-consuming again, as large and salient object classes are already annotated.
Indeed, we find that larger objects are typically annotated first: object size has a high median rank correlation with the annotation order (\att{$-0.80$}). Interestingly, on the interface of \cite{lin14eccv}, this effect is less pronounced (\att{$-0.50$}), as the annotation order is affected by the symbol search and grouping of classes in the hierarchy.
Finally, our analysis shows that the annotators spend \att{3.9}s between saying the last class name and submitting the task, indicating that they do a thorough final scan of the image to ensure they do not miss any class.

\begin{figure}[t]
	\centering
	\includegraphics[trim={0 0 0 1.2cm},clip,width=1\linewidth]{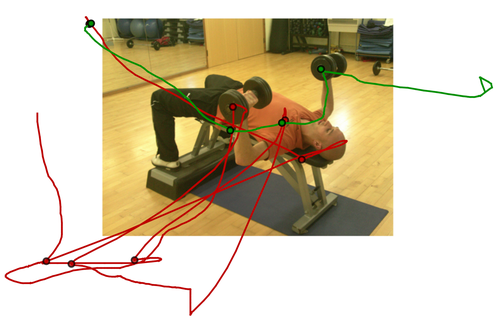}
	\vspace{-0.6cm}
	\caption{A comparison of typical mouse paths produced when annotating an image with our interface (\textcolor{darkgreen}{green}) or with \cite{lin14eccv} (\textcolor{red}{red}). Circles indicate clicks. Mouse paths for our interface are extremely short, thanks to its simplicity and naturalness.}
	\label{fig:mouse_path}
\end{figure}

\para{Mouse path length.}
To better understand the amount of work required to annotate an image we also analyse the mean length of the mouse path. We find that on \ilsvrc{} annotators using \cite{lin14eccv} move the mouse for a \att{$3.0\times$} greater length than annotators using our interface.
Thus, our interface is not only faster in terms of time, but is also more efficient in terms of mouse movements.
The reason is that the hierarchical interface requires moving the mouse back and forth between the image and the class hierarchy (Fig.~\ref{fig:mouse_path}).
The shorter mouse path indicates the simplicity and improved ease of use of our interface.

\para{\cam{Training time.}}
Training annotators to achieve good performance on the 200 classes of~\ilsvrc{} takes 1.6 hours for our interface, or 1 hour with the hierarchical interface of \cite{lin14eccv}.
Instead, annotating the full~\ilsvrc{} dataset would take \num{1726} hours with our interface \vs \num{4474} hours with \cite{lin14eccv}.
Hence, the cost of training is negligible and our interface is far more efficient than \cite{lin14eccv} even after
taking training into account.

\para{Transcription accuracy.}
The annotator training task produces spoken and written class names for each annotated object (Sec.~\ref{sec:training}).
Using this data we evaluate the accuracy of the automatic speech recognition (ASR). For this we only take objects into account if they have transcriptions results attached.
This keeps the analysis focused on transcription accuracy by ignoring other sources of errors, such as incorrect temporal alignment or annotators simply forgetting to say the class name
after they click on an object.

Tab.~\ref{tab:transcription_accuracy} shows the transcription accuracy in two setups: with and without using the vocabulary as phrase hints.
Phrase hints allow to indicate phrases or words that are likely to be present in the speech and thus help
the ASR model transcribe them correctly more often.
Using phrase hints is necessary to obtain high transcription accuracy.
Thanks to them, Recall@3 is at \att{96.5\%} on \coco{} and \att{97.5\%} on \ilsvrc{}.
Hence, the top three transcriptions usually contain the correct class name, which we then extract as described in Sec.~\ref{sec:speech_to_class}.

In fact, we actually consider the above numbers to be a lower bound on the transcription accuracy in the main task, as here we compare the transcriptions against the raw written class names, which contain a few spelling mistakes.
Moreover, here the annotators are in the training phase and hence still learning about the task.
Overall, the above evidence shows that ASR achieves high accuracy, definitely good enough for labelling object class names.

\para{Vocabulary usage.}
As speech is naturally free-form, we are interested in knowing how often annotators use object names that are outside of the vocabulary.
Thus, we analyse how often the written class name in the annotator training task does not match a vocabulary name.
We find that on \coco{} annotators are essentially only using names from the vocabulary (\att{99.5\%} of the cases).
On \ilsvrc{} they still mostly use names from the vocabulary, despite the greater number of classes which induces a greater risk of misremembering their names (\att{96.3\%} are in vocabulary).

Some of the out-of-vocabulary names are in fact variations of names in the vocabulary. These cases can be mapped to their correct name in the vocabulary as described in Sec.~\ref{sec:speech_to_class}.
For example, for the~\ilsvrc{} dataset some annotators say ``oven'', which gets correctly mapped to ``stove'', and ``traffic signal'' to ``traffic light''.
In other cases the annotators use out-of-vocabulary names because they actually label object classes that are not in the vocabulary (\eg ``fork'' and ``rat'', which are not classes of~\ilsvrc{}).

We find that our annotator training task helps reducing the use of out-of-vocabulary names: on \ilsvrc{} the use of vocabulary names increases from 96.3\% in training to 97.5\% in the main task.

\begin{table}
	\centering
	\resizebox{1\linewidth}{!}{%
	\setlength{\tabcolsep}{12pt}
	\begin{tabular}{|c|r|r|}
		\hline
		&  \textbf{Recall@1} & \textbf{Recall@3} \\
		\hline
		\coco{} w/ hints & 93.1 \% & 96.5 \% \\
		\coco{} w/o hints & 70.5 \% & 84.7 \% \\
		\hdashline
		\ilsvrc{} w/ hints & 93.3 \% & 97.5 \%  \\
		\ilsvrc{} w/o hints & 70.2 \% & 89.5 \%  \\
		\hline
	\end{tabular}}
	\caption{\textbf{Transcription accuracy.} Accuracy is high when using phrase hints (see text).
	}
	\label{tab:transcription_accuracy}
\end{table}

\begin{figure}[t]
	\centering\includegraphics[width=1\linewidth]{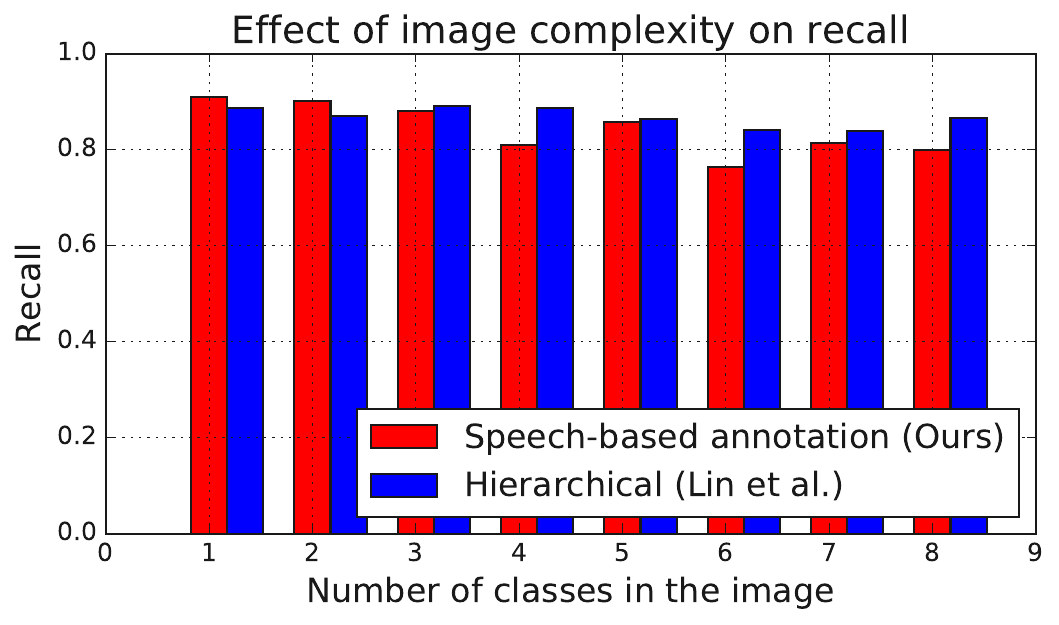}
	\caption{\changedtext{\textbf{Recall as a function of the number of distinct object classes present in an image on the COCO dataset}.
	We find that for cluttered images, annotators tend to miss fewer objects with \cite{lin14eccv} compared to our method.}}
	\label{fig:num_classes_vs_recall}
\end{figure}

\changedtext{
\para{Error analysis.} To better understand the limits of our method we conducted a
detailed analysis of the errors annotators make. We analyzed the recall per
class and as a function of the number of classes in an image.

In terms of average recall per class we find no significant difference between the annotations produced by the hierarchical interface of \cite{lin14eccv} and our method on the COCO dataset (\attc{82.1\%} vs \attc{82.8\%}).
On the ILSVRC dataset, our method has a somewhat lower average class recall of \attc{76.7\%} compared to \cite{lin14eccv} with \attc{84.5\%}.
This is consistent with the difference in the overall recall (Tab.~\ref{tab:accuracy}).
We attribute this difference to the increased challenge of spotting 200 different classes, without being explicitly asked about their presence.

Secondly, we analyzed recall as a function of the number of distinct object classes present in an image.
On the COCO dataset, our method delivers equal or slightly better recall to the hierarchical approach for images with up to 3 classes (which is the most common case, Fig. \ref{fig:num_classes_vs_recall}).
For cluttered images with more than 3 classes, the recall of our method decreases to slightly below that of the hierarchical approach.
In such images objects are often small and hard to spot, hence explicitly querying for then, as done in \cite{lin14eccv}, can help to find such objects.
On ILSVRC, which consists of simpler images with fewer distinct classes in an image (\attc{1.6} in average), this effect is less visible.
}

\section{Experiments on Bounding Box Annotation}
\label{sec:experiments_xclick}

We now present results on using speech to simultaneously annotate objects with a bounding box and their class label.
Thereby we compare our approach (Sec.~\ref{sec:xclick_task}) to a standard two-stage approach.
As in the previous experiment we use 5 crowd workers, which each annotate the 80 classes of the \coco{} dataset on 300 images.

Below we start by briefly explaining the two-stage baseline (Sec.~\ref{sec:two_stage}), before presenting the results of our method (Sec.~\ref{sec:joint_results}).
Finally, Sec.~\ref{sec:joint_additional_analysis} provides additional analysis of our interface.

\subsection{Two-stage approach}
\label{sec:two_stage}

We evaluate the standard way to annotate images with object bounding boxes \cite{russakovsky15ijcv,kuznetsova18arxiv,su12aaai} or outlines~\cite{lin14eccv}, which is typically done in two stages.
In the first stage, annotators are asked to mark the presence or absence of object classes in each image.
For this, we use the results produced by the hierarchical interface of \cite{lin14eccv} (Sec.~\ref{sec:experiments_baseline}).
Thereby we use the object class labels produced by a single annotator, randomly chosen for each image.
In the second stage, annotators are given one of these class labels and are asked to draw bounding boxes for all objects of that class.
For this, we use the efficient extreme clicking interface \cite{papadopoulos17iccv}.
This task is repeated for each class marked as present.
We use the same interface as the one presented in Sec.~\ref{sec:joint_task}, adapted to this task.

\para{Results.}
Tab.~\ref{tab:xclick_results} shows results.
The first stage of the two-stage approach takes \att{29.9} seconds per image (Sec.~\ref{sec:experiments_baseline}).
Then, the second stage takes \att{7.4} seconds per box \cite{papadopoulos17iccv, kuznetsova18arxiv}.
We can estimate the total cost per box by diving the cost of the first stage by the average number of boxes per image (\att{5.1}s), and then adding the cost of the second stage (7.4s). This gives a total cost of \att{12.5} seconds per box.

Furthermore, as a sanity check we evaluate if the bounding boxes produced in this experiment are semantically correct, by comparing their class labels against the ground truth.
Specifically, for each annotated bounding box, we find the ground truth box with the highest overlap and check whether the two boxes
have the same label. We ignore annotated boxes for which there is no corresponding ground truth box. %
We find that the resulting boxes have high semantic accuracy, with \att{96.4\%} of the classes being correct.
We further evaluate geometrical accuracy using mean intersection-over-union (IoU). %
The bounding boxes have a mean IoU of \att{84.4\%}, close to the human agreement upper-bound of 88\%~\cite{papadopoulos17iccv,kuznetsova18arxiv}.
Hence, we conclude that the data produced by this baseline experiment is of high quality.

\subsection{Results for Simultaneous Class and Box Labelling}
\label{sec:joint_results}

We analyze the time per box for our method in Tab.~\ref{tab:xclick_results}.
Simultaneously annotating one object with a bounding box and class label with our method takes \att{6.5 seconds} on average.
It thus provides a significant speedup of \att{$1.9\times$} over the two-stage approach.

This experiment shows the power of speech as annotation modality, as the class label can in fact be annotated at \emph{zero additional cost} over just annotating bounding boxes.
This is because speaking and pointing can be done in parallel \cite{kahneman73attention, oviatt03book}.
Interestingly, we find the joint approach to be even slightly faster than bounding box annotation alone. %
This may seem surprising, but it can be explained by how the objects are annotated in the two cases:
In our approach, all classes are annotated at once. Hence, the annotator parses the image only once, actively searching for all objects across all classes in the vocabulary.
In the two-stage approach instead, annotators draw bounding boxes of each class separately. %
For each class the image is presented again, hence requiring repeated visual search.
This small extra cost translates into about $0.9s$ per box on average.
While it is not an intrinsic advantage of using speech, as the two-stage approach could be reorganized with a smarter interface that asks to box all classes at once, we believe it is an interesting effect.

\begin{figure*}[t]
  \centering
  \resizebox{1\linewidth}{!}{%
  \centering\includegraphics[height=0.22\linewidth,trim={0 0 0 0},clip]{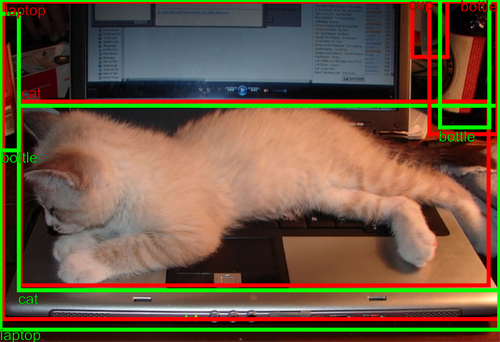}%
  \centering\includegraphics[height=0.22\linewidth,trim={0 0 0 0},clip]{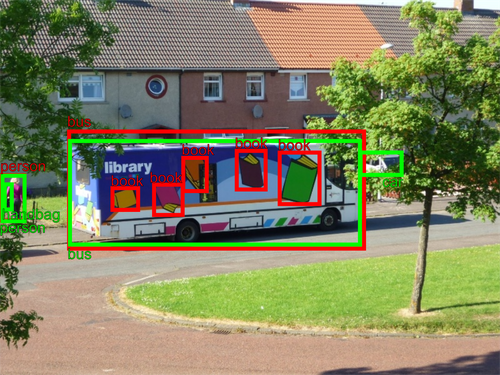}%
  \centering\includegraphics[height=0.22\linewidth,trim={0 0 0 0},clip]{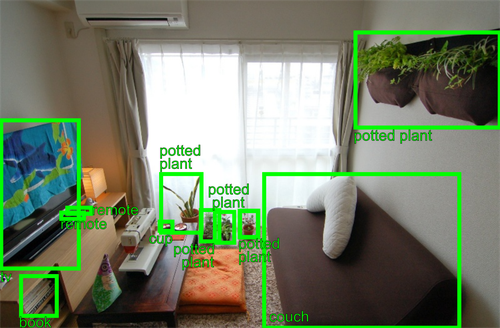}%
  }
  \caption{\changedtext{\textbf{Example annotations produced with simultaneous \jointtask{} on COCO.}
  Original annotations in \textbf{\textcolor{red}{red}}, annotations produced with our method in \textbf{\textcolor{green}{green}}.
  Our method produces accurate bounding boxes and semantic labels.
  In some cases, annotators delimit object classes differently than in the original annotation (center).
  Annotators are able to simultaneously annotate semantic labels and boxes, even for complex images with many objects present (right, original annotations omitted for better visibility).}
  }

  \label{fig:examples_xclick}
\end{figure*}

In terms of quality, we find that our approach produces bounding boxes with semantically accurate class labels (\att{94.2\%}).
This is, however, slightly below the two-stage approach (\att{$-2.2$}\%),
which can be attributed to transcription and alignment errors.
Geometrical accuracy is high and similar to the two-stage approach (\att{83.4\%} \vs \att{84.4}\%).
Thus, we conclude that our method produces annotations with comparable accuracy, but at a significant speed gain of \att{$1.9\times$}.
\changedtext{
We show example annotations generated with our interface in Fig.~\ref{fig:examples_xclick}.
}

\begin{table}[t]
  \centering
  \resizebox{1\linewidth}{!}{%
  \setlength{\tabcolsep}{6pt}
  \renewcommand{\arraystretch}{1.2}%
  \begin{tabular}{|c|c|c|}
    \hline
    &  \parbox{1.7cm}{\textbf{Ours (Box \& Speak)}}& \parbox{1.7cm}{\textbf{Two-stage approach}} \\
    \hline
    Semantic accuracy & 94.2\% & 96.4\% \\
    IoU & 83.4\% & 84.4\% \\
    \hline
    Time / box & \att{6.5} sec. & \att{12.5} sec. \\
    & & (\att{5.1}s + \att{7.4}s) \\
    \hline
  \end{tabular}
  }
  \caption{\textbf{Bounding Box annotation results.}
  Our method of jointly providing class labels and boxes is \att{$1.9\times$} faster than
  the standard two-stage approach. Dataset: COCO.
  }
  \label{tab:xclick_results}
\end{table}

\subsection{Additional analysis of our interface}
\label{sec:joint_additional_analysis}

\begin{figure}[t]
  \centering\includegraphics[width=1\linewidth]{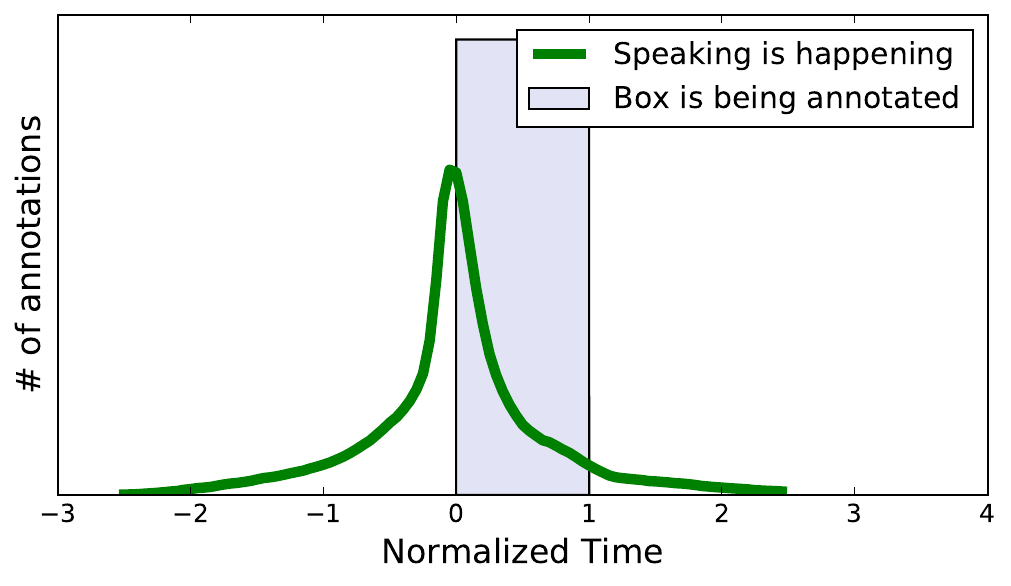}
  \vspace{-0.2cm}
  \caption{
  {\bf Synchronicity of speaking and box drawing.}
  The green curve shows the number of objects (vertical axis) for which the annotator was speaking at a particular point in time (horizontal axis).
  The horizontal axis is normalized by the amount of time needed to draw the box for a particular object.
  This curve shows that people typically start speaking a little before the do the first click on the object (time point at 0) and finish speaking mostly by the time they did the last click on the object (time point at 1).}
  \label{fig:xclick_voice_overlap}
\end{figure}

\para{Concurrency of speaking and clicking.}
\changedtext{
In Fig.~\ref{fig:xclick_voice_overlap}
we analyze the relative time at which annotators provide bounding boxes and say the class name.
We find that annotators typically do both at roughly the same time, but have a tendency to start speaking before clicking on the object.
This matches previous studies, which found that annotators show multiple patterns of multimodal annotation, where one input often partially precedes the other \cite{oviatt97integration, oviatt03book}.
In fact, we observed this effect despite instructing annotators to ``mark the object and speak at the same time''.
We conjecture that annotators start speaking after spotting an object and while they move the mouse to the first click position.
In fact this tendency is occasionally so strong that there is no temporal overlap between saying the class name and marking the object location (3.6\% of the cases).

As a consequence of this variability, correctly aligning bounding boxes and class names is not trivial.
One potential solution to this challenge could be to assume a fixed pattern and force annotators to use it.
As an example: we could only transcribe the speech between the first and last click of annotating a bounding box. However, this will lead to deriving the class name from incomplete audio when the annotator does not strictly follow this imposed pattern.
Hence, we opt to let the annotators behave naturally, without enforcing a fixed annotation pattern. Instead we handle the resulting differences between the time they speak and point via our robust alignment method
(Sec.~\ref{sec:postprocessing})
which we evaluate next.
}

\para{Comparison of alignment methods.}
While our previous paper \cite{gygli19cvpr} used a heuristic method for temporal segmentation and alignment,
this work proposes a principled approach to align the class labels and object location annotations (Sec.~\ref{sec:postprocessing}).
Using our method leads to class labels with a semantic accuracy
of \att{94.2\%}, compared to \att{80.9\%} for \cite{gygli19cvpr}.
We also find that some boxes have no class label attached,
due to issues in speech recognition, alignment errors or annotators forgetting to say the class name.
For our alignment method, \att{1.2\%} of the bounding boxes have no label, compared to \att{10.8\%} for \cite{gygli19cvpr}.
\changedtext{
Hence, our method is not only more principled, it also minimizes alignment errors,
which leads to considerably better accuracy in practice.
At the same time it remains fast. Inferring class names and aligning them with the bounding box annotations takes $\approx 0.025s$	 per image on average, on both ILSVRC and COCO.
}

\section{Discussion}
\label{sec:discussion}

  \changedtext{
  Based on our experiments and observations we now discuss some of the insights on the advantages,
  limitations and open questions in using speech-based annotation.

  \para{Annotating irrelevant images.}
  In order to be able to measure semantic and location accuracy of our method,  we re-annotated existing datasets.
  Thus, the images we used all contain at least one object of the classes contained in the vocabulary.
  We did not evaluate how our method compares to previous methods on images that do not contain contain any objects of the relevant classes.
  However, this case is rare in practice, as images in standard datasets are not uniformly sampled out of the space of all possible images.
  Instead, datasets are typically created by defining a fixed vocabulary of classes and then explicitly retrieving images that contain these classes from web search engines,~\eg COCO \cite{lin14eccv} and ILSVRC \cite{russakovsky15ijcv}.
  More recently, the Open Images dataset was constructed by uniformly sampling images from Flickr, but then annotates a large vocabulary (600 classes) deliberately chosen to cover objects that are frequent and important in these images~\cite{kuznetsova18arxiv}. Hence, essentially all images contain at least one object from the vocabulary.
  Besides, we believe that even on irrelevant images our method would not be slower than hierarchical approaches \cite{lin14eccv,deng14chi}.
  In our method, the annotator would spend their time carefully scanning the whole image looking for any class in the vocabulary.
  In the hierarchical methods instead, the annotator would have to scan the image looking for any class of each top-level supercategory in turn
  (e.g. ``living organisms'', %
  ``carpentry items'' or ``items that run on electricity'' in ILSVRC).
  While the latter might be a simpler task, it needs to be repeated for each supercategory.

  \para{Scaling to larger vocabularies.}
  We experiment with 200 classes, a size which lies in the ballpark of the largest current datasets for object detection.
  Scaling to larger vocabularies is a challenge for any annotation method.
  While our method relies on memorization of the class names for fast annotation, we enable the annotator to quickly review the class vocabulary during annotation (Sec.~\ref{sec:main_task}).
  In practice, annotators rarely depend on this and only 7.2\% of the total annotation time is spend looking up class names for a vocabulary of 200 classes (Sec. ~\ref{sec:detailed_experiments}).
  This suggests that our approach might scale to many more classes as well.
  Importantly, scaling hierarchical methods also has its challenges, most notably the construction of a hierarchical representation of classes that
  is intuitive for annotators to use.
  Even for 200 classes, this is not trivial and influences the final results as noted by \cite{russakovsky15ijcv}.

  Generally, scaling exhaustive annotation to thousands of classes is an unsolved problem and it is unclear how well any existing method would work in practice as no experiments have been reported
  (as opposed to annotating a few classes from a large vocabulary in each image \cite{deng09cvpr}.
  In our experience the main challenge is not memorizing the class names, but teaching annotators how to recognize, distinguish and delimit classes within a large vocabulary.
  For example, in ILSVRC, annotated with a hierarchical interface, lobsters and scorpions are often confused and sometimes annotated as both.
  How to handle this challenge is a topic of ongoing research~\cite{pont2019arxiv}.

  \para{Mapping speech to object classes.}
  In preliminary experiments we made annotators manually select the class name out of the 3 most likely classes according to our method (Sec.~\ref{sec:speech_to_class}).
  However, we found that such a manual selection takes considerable time, reducing the efficiency gains of using speech. Hence, we opted for automatically selecting the most likely class (Sec.~\ref{sec:speech_to_class}).
  This approach is significantly faster while still delivering accurate annotations
(Tab.~\ref{tab:accuracy} \& \ref{tab:transcription_accuracy}),
hence providing a better speed-accuracy trade-off.

  }

\section{Conclusion}
\label{sec:conclusion}

  We use multimodal inputs for fast image annotation.
  At the core of our method lies speech: annotators provide class labels by simply by saying the names of the objects that are present in an image.
  We have proposed two kinds of speech-based interfaces:
  First, an interface for \taskname{}, a task that has traditionally been time consuming and difficult to design.
  We have shown that our method offers considerable speed gains, thanks to speech:
  it is \att{$2.3\times - 14.9\times$} faster than previous methods~\cite{lin14eccv,deng14chi}. %
  Second, an interface for simultaneous \jointtask{}.
  Previous methods annotate the two in separate stages \cite{russakovsky15ijcv,kuznetsova18arxiv,su12aaai}.
  Instead, we have shown that using speech allows to naturally combine them, which makes the overall process
  \att{$1.9{\times}$} faster than previous methods.
  This is thanks to the fact that saying the class name while drawing a bounding box can be done at zero additional cost.
  Finally, we have conducted a detailed analysis of our interfaces, speech transcription and temporal alignment.
  We believe this offers helpful insights for building even more efficient annotations tools in the future.

\bibliographystyle{spbasic}
\bibliography{shortstrings,loco,references}

\newpage
\setcounter{section}{0}
\renewcommand{\thesection}{Appendix \Alph{section}}
\section{- Two-level Hierarchy for \ilsvrc{}}
\label{sec:appendix_a}

For reference we provide the hierarchy we constructed to use the interface of~\cite{lin14eccv} with the 200 class vocabulary of the~\ilsvrc{} dataset~\cite{deng09cvpr}.
The hierarchy is based on the hierarchy of questions supplied in~\cite{deng09cvpr}, but modified to balance the size of the groups and reduced to two-levels.
It consists of 22 semantic groups and a small group of ``misc objects'':
\small
\begin{enumerate}
\item    Wind instruments:
  \\ \begin{enumerate*}[noitemsep,topsep=0pt,itemjoin={;\quad}]
 \item trumpet
      \item saxophone
      \item trombone
      \item flute
      \item oboe
      \item harmonica
      \item french horn
      \item accordion
   \end{enumerate*}
   
 \item Other musical instruments:
  \\ \begin{enumerate*}[noitemsep,topsep=0pt,itemjoin={;\quad}]
 \item piano
      \item guitar
      \item violin
      \item chime
      \item maraca
      \item drum
      \item cello
      \item banjo
      \item harp
   \end{enumerate*}
   
 \item Fruit:
  \\ \begin{enumerate*}[noitemsep,topsep=0pt,itemjoin={;\quad}]
 \item pineapple
      \item fig
      \item orange
      \item banana
      \item strawberry
      \item apple
      \item lemon
      \item pomegranate
   \end{enumerate*}
   
 \item Other food:
  \\ \begin{enumerate*}[noitemsep,topsep=0pt,itemjoin={;\quad}]
 \item pizza
      \item guacamole
      \item popsicle
      \item hamburger
      \item hotdog
      \item burrito
      \item pretzel
      \item mushroom
      \item bagel
      \item artichoke
      \item cucumber
      \item bell pepper
      \item cabbage
   \end{enumerate*}
   
 \item Clothing:
  \\ \begin{enumerate*}[noitemsep,topsep=0pt,itemjoin={;\quad}]
 \item miniskirt
      \item diaper
      \item brassiere
      \item bathing cap
      \item bow tie
      \item helmet
      \item tie
      \item swimming trunks
      \item swimsuit
      \item hat
      \item sunglasses
   \end{enumerate*}
   
 \item Flying Animals:
  \\ \begin{enumerate*}[noitemsep,topsep=0pt,itemjoin={;\quad}]
 \item bee
      \item ladybug
      \item butterfly
      \item dragonfly
      \item bird
   \end{enumerate*}
   
 \item Felines and Canines:
  \\ \begin{enumerate*}[noitemsep,topsep=0pt,itemjoin={;\quad}]
 \item tiger
      \item lion
      \item domestic cat
      \item fox
      \item dog
   \end{enumerate*}
   
 \item 
    Animals with hooves:
  \\ \begin{enumerate*}[noitemsep,topsep=0pt,itemjoin={;\quad}]
 \item camel
      \item hippopotamus
      \item swine
      \item cattle
      \item zebra
      \item sheep
      \item horse
      \item antelope
   \end{enumerate*}
   
 \item Animals with 6 or more legs:
  \\ \begin{enumerate*}[noitemsep,topsep=0pt,itemjoin={;\quad}]
 \item lobster
      \item scorpion
      \item isopod
      \item centipede
      \item ant
      \item tick
   \end{enumerate*}
   
 \item Animals with no legs:
  \\ \begin{enumerate*}[noitemsep,topsep=0pt,itemjoin={;\quad}]
 \item snake
      \item goldfish
      \item jellyfish
      \item ray
      \item snail
      \item starfish
      \item whale
      \item seal
   \end{enumerate*}
   
 \item Other animals:
  \\ \begin{enumerate*}[noitemsep,topsep=0pt,itemjoin={;\quad}]
 \item red panda
      \item porcupine
      \item giant panda
      \item rabbit
      \item koala
      \item elephant
      \item otter
      \item squirrel
      \item monkey
      \item hamster
      \item skunk
      \item armadillo
      \item bear
      \item frog
      \item lizard
      \item turtle
   \end{enumerate*}
   
 \item Vehicles:
  \\ \begin{enumerate*}[noitemsep,topsep=0pt,itemjoin={;\quad}]
 \item airplane
      \item golfcart
      \item watercraft
      \item train
      \item bus
      \item snowmobile
      \item bicycle
      \item unicycle
      \item snowplow
      \item car
      \item motorcycle
      \item cart
   \end{enumerate*}
   
 \item Cosmetics:
  \\ \begin{enumerate*}[noitemsep,topsep=0pt,itemjoin={;\quad}]
 \item lipstick
      \item face powder
      \item perfume
      \item hair spray
      \item cream
   \end{enumerate*}
   
 \item Medical items:
  \\ \begin{enumerate*}[noitemsep,topsep=0pt,itemjoin={;\quad}]
 \item neck brace
      \item stethoscope
      \item band aid
      \item syringe
      \item stretcher
      \item crutch
   \end{enumerate*}
   
 \item Furniture:
  \\ \begin{enumerate*}[noitemsep,topsep=0pt,itemjoin={;\quad}]
 \item bench
      \item chair
      \item bookshelf
      \item babys bed
      \item table
      \item sofa
      \item filing cabinet
   \end{enumerate*}
   
 \item Carpentry items:
  \\ \begin{enumerate*}[noitemsep,topsep=0pt,itemjoin={;\quad}]
 \item axe
      \item nail
      \item power drill
      \item chain saw
      \item screwdriver
      \item hammer
   \end{enumerate*}
   
 \item School supplies:
  \\ \begin{enumerate*}[noitemsep,topsep=0pt,itemjoin={;\quad}]
 \item pencil box
      \item pencil sharpener
      \item rubber eraser
      \item ruler
      \item binder
   \end{enumerate*}
   
 \item Game equipment:
  \\ \begin{enumerate*}[noitemsep,topsep=0pt,itemjoin={;\quad}]
 \item baseball
      \item golf ball
      \item tennis ball
      \item racket
      \item rugby ball
      \item volleyball
      \item ping-pong ball
      \item croquet ball
      \item basketball
      \item soccer ball
      \item puck
   \end{enumerate*}
   
 \item Sports equipment:
  \\ \begin{enumerate*}[noitemsep,topsep=0pt,itemjoin={;\quad}]
 \item dumbbell
      \item balance beam
      \item horizontal bar
      \item ski
      \item bow
      \item punching bag
   \end{enumerate*}
   
 \item Consumer electronics:
  \\ \begin{enumerate*}[noitemsep,topsep=0pt,itemjoin={;\quad}]
 \item remote
      \item digital clock
      \item computer mouse
      \item computer keypad
      \item laptop
      \item printer
      \item iPod
      \item screen
      \item tape player
      \item microphone
   \end{enumerate*}
   
 \item Electronic appliances:
  \\ \begin{enumerate*}[noitemsep,topsep=0pt,itemjoin={;\quad}]
 \item washer
      \item coffee maker
      \item microwave
      \item waffle iron
      \item toaster
      \item refrigerator
      \item stove
      \item dishwasher
      \item vacuum
      \item electric fan
      \item hair drier
   \end{enumerate*}
   
 \item Non-electric kitchen items:
  \\ \begin{enumerate*}[noitemsep,topsep=0pt,itemjoin={;\quad}]
 \item bowl
      \item ladle
      \item salt shaker
      \item can opener
      \item cocktail shaker
      \item frying pan
      \item spatula
      \item plate rack
      \item strainer
      \item corkscrew
      \item water bottle
      \item mug
      \item pitcher
      \item wine bottle
      \item milk can      
   \end{enumerate*}
   
 \item Misc objects:
  \\ \begin{enumerate*}[noitemsep,topsep=0pt,itemjoin={;\quad}]
 \item person
      \item traffic light
      \item flowerpot
      \item purse
      \item backpack
      \item plastic bag
      \item lamp
      \item beaker
      \item soap dispenser      
   \end{enumerate*}
   \end{enumerate}

\end{document}